\documentclass[letterpaper, 10 pt, conference]{ieeeconf}  

\IEEEoverridecommandlockouts                              

\title{\LARGE \bf CoNVOI: Context-aware Navigation using Vision Language Models in Outdoor and Indoor Environments
}

\author{Adarsh Jagan Sathyamoorthy$^{1}$, Kasun Weerakoon$^{1}$, Mohamed Elnoor$^{1}$, Anuj Zore$^{1}$, \\ Brian Ichter$^{2}$, Fei Xia$^{2}$, Jie Tan$^{2}$, Wenhao Yu$^{2}$, and Dinesh Manocha$^{1}$ \\
{\small Technical report, code, and video can be found at \href{http://gamma.umd.edu/convoi/}{http://gamma.umd.edu/convoi/}}
\thanks{$^{1}$ Authors are with the University of Maryland, College Park.}
\thanks{$^{2}$ Authors are with Google DeepMind.}
}

\usepackage{xcolor}
\usepackage{amssymb}
\usepackage{graphicx}
\usepackage{mathtools}
\usepackage{amsmath}
\usepackage{gensymb}
\usepackage{float}
\usepackage{multirow}
\usepackage{makecell}
\usepackage{mathtools}
\usepackage{hyperref}
\hypersetup{
    colorlinks=true,
    linkcolor=blue,
    filecolor=magenta,      
    urlcolor=blue,
    pdftitle={Overleaf Example},
    pdfpagemode=FullScreen,
    }

\usepackage{tabularx}
    \newcolumntype{L}{>{\raggedright\arraybackslash}X}
\usepackage[utf8]{inputenc}
\usepackage[english]{babel}

\usepackage{amsthm}

\usepackage{algorithm}
\usepackage{algpseudocode}

\usepackage{subcaption}

\newcommand{\ours}{CoNVOI}
\newcommand{\no}{\noindent}

\let\emptyset\varnothing
\begin{document}

\maketitle
\thispagestyle{empty}
\pagestyle{empty}


\begin{abstract}
We present \ours, a novel method for autonomous robot navigation in real-world indoor and outdoor environments using Vision Language Models (VLMs). We employ VLMs in two ways: first, we leverage their zero-shot image classification capability to identify the \textit{context} or scenario (e.g., indoor corridor, outdoor terrain, crosswalk, etc) of the robot's surroundings, and formulate context-based navigation behaviors as simple text prompts (e.g. ``\textit{stay on the pavement}"). Second, we utilize their state-of-the-art semantic understanding and logical reasoning capabilities to compute a suitable trajectory given the identified context. To this end, we propose a novel multi-modal visual marking approach to annotate the obstacle-free regions in the RGB image used as input to the VLM with numbers, by correlating it with a local occupancy map of the environment. The marked numbers ground image locations in the real-world, direct the VLM's attention solely to navigable locations, and elucidate the spatial relationships between them and terrains depicted in the image to the VLM. Next, we query the VLM to select numbers on the marked image that satisfy the context-based behavior text prompt, and construct a reference path using the selected numbers. Finally, we propose a method to extrapolate the reference trajectory when the robot's environmental context has not changed to prevent unnecessary VLM queries. We use the reference trajectory to guide a motion planner, and demonstrate that it leads to human-like behaviors (e.g. not cutting through a group of people, using crosswalks, etc.) in various real-world indoor and outdoor scenarios. We perform several ablations and navigation comparisons and demonstrate that \ours's trajectories are most similar to human teleoperated ground truth in terms of Fréchet distance (9.7-58.2\% closer), lowest path errors (up to 88.13\% lower), and up to 86.09\% lower \% of unacceptable paths.
\end{abstract}

\section{Introduction}

Wheeled and legged robots have been used to navigate many kinds of challenging indoor and outdoor environments for applications such as delivery, surveillance, job-site monitoring in construction, disaster response, etc. The navigation behaviors required while performing these tasks can vary significantly in indoor and outdoor scenarios. For instance, in indoor settings, a robot must exhibit socially acceptable behaviors in narrow corridors, and in the presence of humans. In outdoor environments, the robot must avoid navigating on uneven or bumpy terrains and roads with oncoming traffic. It should favor stable, paved surfaces, identify locations to cross streets/roads, etc.

\begin{figure}[t]
    \centering
    \includegraphics[width=0.9\columnwidth]{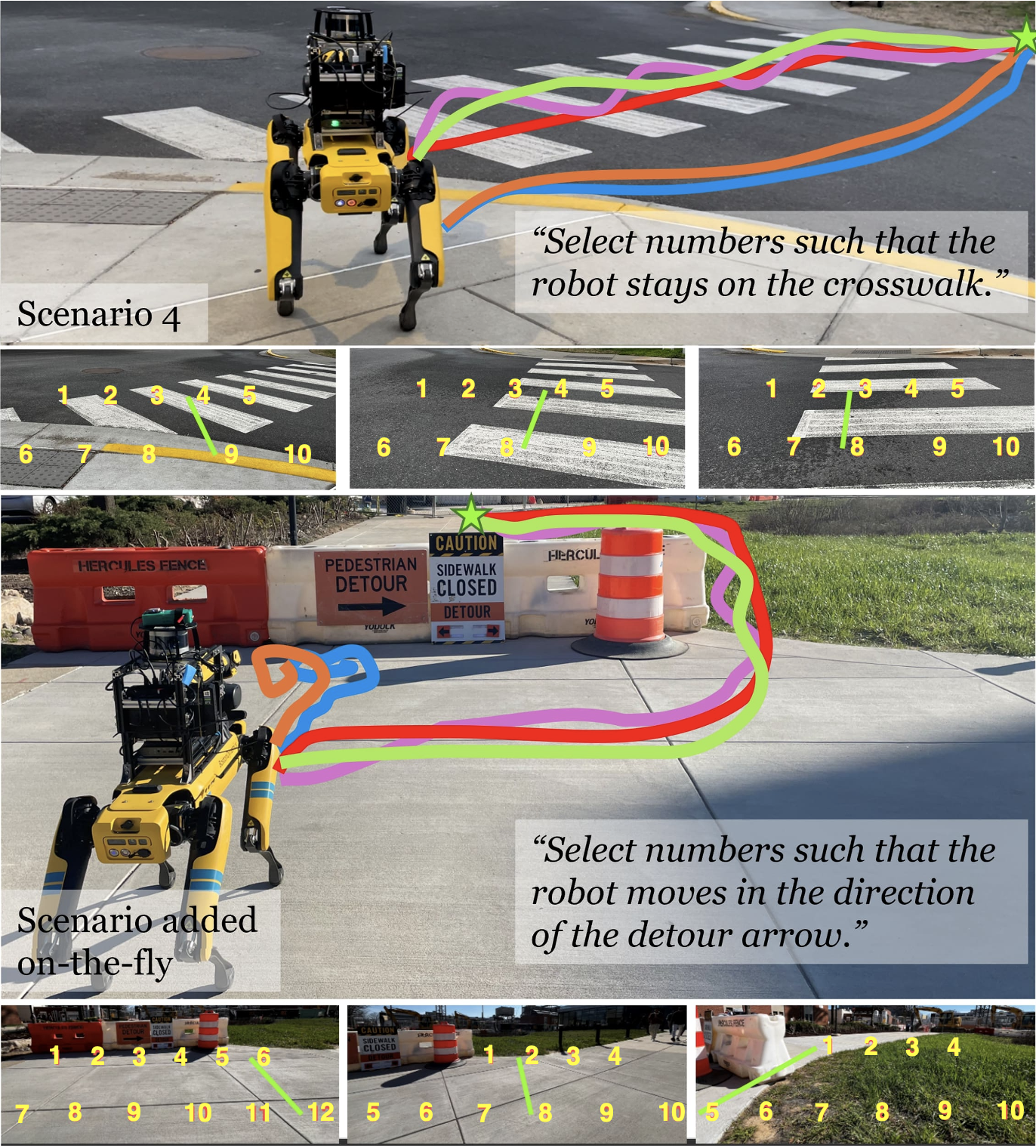}
    \caption{\small{\textbf{[Top]}: The trajectories of a Spot robot crossing the road when using \ours \, with GPT-4v \cite{gpt-4v} (in green), \ours \, with Gemini \cite{gemini} (in purple), teleoperated by a human (in red), GA-Nav \cite{guan2021ganav}, and DWA \cite{DWA}. \ours \, navigates the robot on the crosswalk by understanding the environmental context. \textbf{[Bottom]}: The trajectories when the Spot robot navigates to a goal beyond a blockade. While using \ours, handling such scenarios can be added on the fly using a simple text prompt without any reformulation. The RGB images from the robot are shown with \ours's multi-modal visual marking (numbers in yellow). GPT-4v/Gemini is queried with the marked image and a context-based text prompt in quotes, and it returns the green reference path that follows explicit, and implicit social rules, and human-like preferences during indoor and outdoor navigation.}}
    \label{fig:cover_image}
    \vspace{-10pt}
\end{figure}

Humans are able to effectively navigate all these environments because we identify the surrounding scenario or \textit{context} (e.g. indoor hallway, outdoor terrain, etc), and follow certain explicit (e.g. crossing roads at crosswalks) and implicit (e.g. keep to one side of a corridor) social rules, and preferences (e.g. walk on well-paved surfaces). Numerous research works in indoor and outdoor navigation have been proposed to exhibit these behaviors. For instance, works in indoor navigation have focused on socially-compliant navigation behaviors in avoiding and overtaking dynamic pedestrians \cite{social-forces,socially-aware-Jon-How,frozone,MPDM}, groups of people \cite{comet-ieee}, keeping to one side in corridors which helps avoid oncoming humans \cite{efficient-nav-rob-humans,human-friendly}, etc. In outdoor navigation, works have predominantly focused on estimating terrain traversability using semantic segmentation \cite{semantic-segmnt,guan2021ganav,shaban2022semantic}, self-supervised learning \cite{terrapn,model-error}, etc, and detecting complex outdoor obstacles \cite{vern}. 
 

However, prior model-based and learning-based approaches are tailored for specific perception or navigation tasks \cite{foundation-models-survey}, limiting their applicability to other challenges or their generalization to different indoor and outdoor settings. Conversely, Large Language Models (LLMs) \cite{gpt-3,gpt-4v,llama2,llava,gemini} that use only text inputs and subsequently large Vision Language Models (VLMs) that use both RGB images and text prompts as inputs \cite{gpt-4v,gemini,llama2,visual-instruction-tuning} have overcome this generalization limitation to a large extent. They have demonstrated impressive zero-shot classification \cite{clip,clip-fields}, semantic visual understanding \cite{ha2022semantic_w_vlm}, and logical reasoning capabilities \cite{pivot} in various tasks involving embodied vision language navigation \cite{clipnav,shah2022lmnav}, open vocabulary manipulation \cite{okrobot,saycan}, etc. These capabilities would be beneficial for perceiving and reasoning about complex indoor and outdoor environments during autonomous navigation. 

However, there are a few key challenges in using large VLMs for navigating mobile robots. Due to their large model size and high memory and computational demands, these models cannot be run on robot-mounted edge processors. They typically require several seconds to process queries and the response time varies based on the size of the queries \cite{chang2023llm_survey}. Therefore, current VLMs cannot be queried for safety-critical tasks such as realtime collision avoidance with dynamic obstacles. Additionally, queries should be succinct and account for the environment's context to elicit quick and accurate responses. 


\textbf{Main Contributions:} We present a novel method to harness the state-of-the-art perception capabilities (generalization, semantic understanding, etc.) of VLMs, obtain a context-based reference path, and integrate it with a motion planning algorithm for real-time robot navigation in complex indoor and outdoor environments. The novel components of are work include:

\begin{itemize}
    \item A novel context-based prompting method that involves querying a compact VLM \cite{clip} to determine the robot's environment context (e.g., indoor corridors, outdoor terrains, crosswalks). Next, our approach utilizes a predefined text prompt that describes a context-based navigation behavior and a marker-overlayed RGB image to a large VLM \cite{gpt-4v,gemini,llava} to generate a reference trajectory for the robot. We demonstrate that using our context-based text prompts helps generate reference trajectories that adhere to explicit and implicit social rules while also mimicking human preferences (e.g. walking on pavement, crossing at crosswalks, etc). 

    \item A novel multi-modal visual marking method that enhances the RGB image inputs to a large VLM by overlaying visual markers/numbering on obstacle-free regions, aiding the VLM to focus on navigable regions, comprehending spatial relationships and underlying terrains across different regions within the image. Our method correlates an occupancy grid map obtained from lidar point clouds with the RGB image to determine the obstacle-free locations in images to add markers. We demonstrate that our visual prompting method results in more accurate responses (up to 88.13\% lower path errors, and preventing unsafe paths up to 86.09\% fewer times) to queries from the VLM when compared with visual markers that are overlayed on obstacles as well. 

    \item A method to extrapolate the reference trajectory to future time steps when the robot's context has not changed. This allows our method to navigate without unnecessarily querying the VLM to generate new reference trajectories. We integrate our VLM-based perception with a planner to demonstrate smooth, uninterrupted, real-time navigation in several indoor and outdoor environments using a Turtlebot and a Spot robot without any domain specific fine-tuning. \ours's trajectories match human teleoperated trajectories (upto 58.26\% closer Fréchet distance) than existing baseline navigation methods.  
\end{itemize}
\section{Related Works}

\subsection{Indoor Navigation}
Many recent works have focused on navigating in indoor settings in a socially acceptable manner \cite{socially-aware,WB1,frozone,Humanaware-survey}. The objective of social navigation is to compute trajectories that are not only collision-free but also follow certain common norms \cite{Humanaware-survey} such as keeping to one side in a corridor, not interrupting groups of people by moving in-between them \cite{comet-ieee}, etc. A robot must also maintain sufficient distance from humans even if they are closer to inanimate obstacles \cite{socially-aware}. 

Many techniques that are based on reinforcement learning (RL) \cite{JHow1, JiaPan1}, and inverse reinforcement learning (IRL) \cite{kitani,pfeiffer} have been proposed to develop such behaviors. These methods train end-to-end models to avoid collisions, sudden rotatory maneuvers, and generate smooth trajectories. However, all these methods focus on modeling \textit{some} aspects of social navigation and do not possess the general logical reasoning \cite{foundation-models-survey} that VLMs may possess to adapt to a new scenario. Our method achieves several social behaviors using our context-aware prompting without any training or reformulation.


\subsection{Outdoor Navigation}
For navigating outdoor environments, a robot must estimate the traversability of the terrains ahead of the robot, and compute trajectories on the most safe, and smooth terrain. To this end, there have been several works in pixel-wise semantic segmentation to classify a terrain into multiple predefined traversability classes (smooth, rough, bumpy, forbidden, etc.) \cite{guan2021ganav,ttm,geo-visual}. These works are typically trained in a supervised manner using human-annotated image datasets. There have also been unsupervised learning-based works that automatically label terrains by characterizing the robot-terrain interaction using other sensor data such as forces/torques \cite{where-should-i-walk}, vibrations and differences in odometry \cite{terrapn, model-error}, acoustic data \cite{proprioceptive-sensor}, vertical acceleration experienced \cite{multi-sensor-correlation}, and stereo depth \cite{long-range-vision,classifier-ensembles}, etc. 

However, such models often do not generalize to new outdoor environments, and the associated planners may not be applicable to indoor settings. Our proposed approach using VLMs is applicable to both indoor and outdoor settings without requiring any environment-specific dataset or training.

\subsection{Vision Language Models in Navigation}
Over the past few years, LLMs \cite{gpt-3} and VLMs \cite{clip,gpt-4v,gemini,llama2,llava,visual-instruction-tuning} have demonstrated impressive accuracy in tasks such as scene/semantic understanding \cite{ansel}, grounding objects in a scene based on language description, instruction following, code generation \cite{code-as-policies}, etc that are useful for robot manipulation and navigation tasks. For instance, Shah et al. \cite{shah2022lmnav} utilized GPT-3 and CLIP to extract landmark descriptions from a text navigation instruction and ground them in images, respectively, to guide a robot's navigation. Further, Shah et al. \cite{semantic-guesswork} demonstrated how the semantic understanding that large VLMs possess can help bias a robot's exploration to search for a user-defined object/location. Similar to \cite{shah2022lmnav}, Chen et al. \cite{a2nav} utilize LLMs to decompose navigation instructions into a series of actionable tasks that are executed by an action-aware navigation policy.

Zhu et al. \cite{VL-decision-system} propose a system to use VLMs to describe a scene, and an LLM to verify if the scene matches a user-defined target. Visual Language Maps \cite{VL-Maps} demonstrate fusing a VLM descriptions of landmarks with a 3D map that enables the robot to understand spatial references relative to landmark. OK-Robot \cite{okrobot} vision-language representations computed from an environmental scan by VLMs are stored as semantic memory \cite{clip-fields}, and is matched with a language query to move to a desired target object and pick it up in a zero-shot manner. 

Existing research has primarily used VLMs to identify navigation targets (\textit{where} to navigate) \cite{clipnav,shah2022lmnav} and parse natural language commands but has not focused on the \textit{how} (lower-level behaviors such as walking on pavements, not disturbing interacting pedestrians) to navigate robots to their targets. Additionally, while many studies have been evaluated in realistic simulated environments, those tested in real-world settings \cite{shah2022lmnav,okrobot,saycan} have not addressed the use of semantic and contextual understanding of the environment to guide the robot's navigation behaviors. \ours's goal is to leverage VLMs to guide low-level navigation behaviors to be the most suitable for a given scenario.


\section{Preliminaries}
In this section, we introduce and define the symbols and key concepts used in our work.

\subsection{Symbols, Definitions, and Assumptions}
\ours \, addresses the challenge ``what navigation behaviors should a robot exhibit in its current local environmental context". This involves determining the locations within the robot's field of view to navigate toward to exhibit these behaviors, extending beyond simple obstacle avoidance.

For our formulation, we assume a robot equipped with an RGB camera and a 2D or 3D lidar rigidly attached to a wheeled or legged mobile robot base. The camera provides an RGB image $I^{RGB}_t$ to view the environment and the lidar provides 2D laser scans at time instant $t$. Our formulation uses an $n \times n$ robot-centric occupancy grid map $\mathcal{O}_t$ obtained using the lidar's laser scan to represent the environmental obstacles. The grids of $\mathcal{O}_t$ are marked with 1's at grids that contain obstacles, and 0's otherwise. The obstacles are inflated by the robot's radius, and the robot is considered as a point at $(n/2, n/2)$, and the set of obstacles is denoted as $obs_{t}$. The pixels in $I^{RGB}_t$ are correlated with the grids in $\mathcal{O}_t$ using a homography perspective projection.

\begin{figure*}[t]
    \centering
    \includegraphics[width=0.9\linewidth]{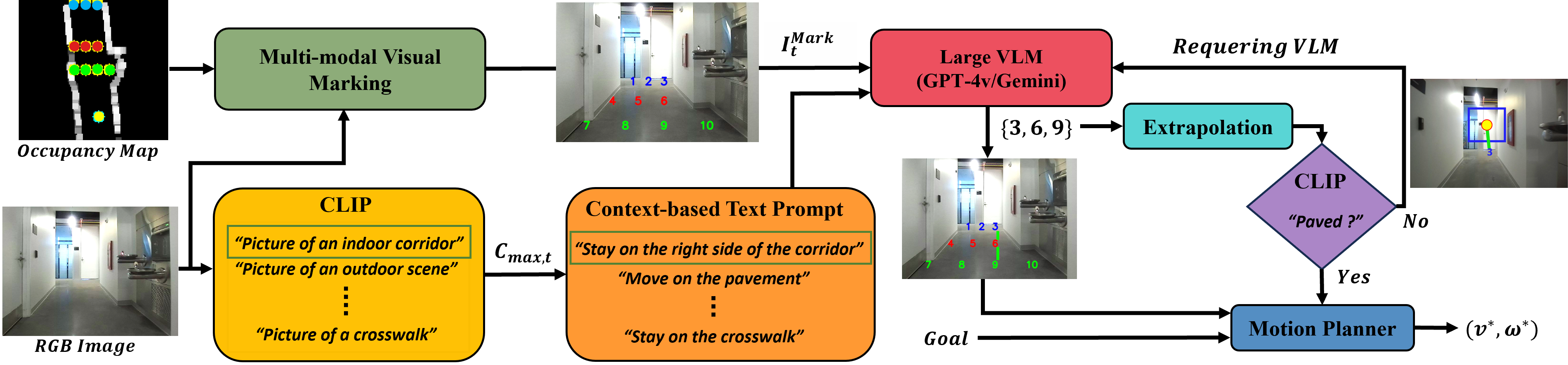}
    \caption{\small{\ours's architecture utilizes CLIP to interpret the context of the robot's environment from an RGB image ($I^{RGB}_t$), identifying features such as indoor corridors, social scenarios with people, outdoor terrains, etc. Next, \ours \, queries a large VLM with a context-based text prompt, and the RGB image marked with numbers ($I^{Mark}_t$) in the free space detected in an occupancy grid map to generate a reference path (in green) that adheres to explicit and implicit social rules (e.g., staying on pavement, using crosswalks). A dedicated motion planner then follows this path while avoiding obstacles. To prevent unnecessarily requerying the large VLM, we extrapolate this reference path linearly and check if the extrapolated point (in yellow) lies on a paved path (sidewalk, corridor, etc) to either use it for navigation, or requery the VLM again. Instead of using VLMs for direct robot control or for well-addressed tasks such as goal-reaching and obstacle avoidance, we leverage its context-understanding capabilities to achieve more intricate, zero-shot navigation behaviors using a separate planner.}}
    \label{fig:system-arch}
    \vspace{-10pt}
\end{figure*}

The robot is controlled using linear and angular velocity commands ($v, \omega$), respectively. We consider \textit{three coordinate frames}: 
1. robot - a frame attached to the robot's center of mass with X, Y, and Z axes pointing forward, leftward, and upward relative to the robot, 2. Occupancy map ($\mathcal{O}$) - a frame attached at the center of $\mathcal{O}_t$ with X, and Y axes pointing forward and leftward, and 3. RGB - a frame attached to the top-left corner of the RGB image obtained from the camera, with X, and Y axes pointing rightward (across columns), and downward (across rows). Quantities belong to the frame indicated by their superscripts (e.g. $x^{rob}$ belongs to the robot frame). Transformation matrices between these frames are represented as $T^{odom}_{rob}$, $T^{rob}_{\mathcal{O}}$, $T^{\mathcal{O}}_{RGB}$. These matrices transform points in the subscript frame to the superscript frame. Finally, we use symbols $i, j$ to denote indices. \ours's overall architecture is shown in Fig. \ref{fig:system-arch}. 

\subsection{Motion Planner} \label{sec:planner-background}
Our formulation uses a separate motion planner integrated with the VLM to smoothly navigate the robot in real-time. In this section, we provide a brief overview of the planner's preliminary concepts. 

We use the method proposed in ~\cite{DWA} for motion planning. We represent the robot's actions as linear and angular velocity pairs $(v,\omega)$. The trajectory $traj^{rob}(v, w)$ that a $(v, \omega)$ pair leads to is calculated as a set of positions $\{(x^{rob}_t, y^{rob}_t), (x^{rob}_{t + \Delta t}, y^{rob}_{t + \Delta t}), ..., (x^{rob}_{t + t_{hor}}, y^{rob}_{t + t_{hor}})\}$ relative to the robot. It is also obtained relative to $\mathcal{O}_t$ and represented as $traj^{\mathcal{O}}(v, \omega)$. Here, $\Delta t$ is a time increment, and $t_{hor}$ is a time horizon until which the trajectory is extrapolated. At any time instant $t$, the optimal velocity pair $(v^*,\omega^*)$ to navigate the robot is calculated by minimizing the objective function below,

\vspace{-10pt}
\begin{equation}
\label{eqn:dwa_obj_func}
    Q_1(v,\omega) = \sigma\big(\alpha_1 . head(.) + \alpha_2 . obs(.) + \alpha_3 . vel(.) \big),
\end{equation}

\no $\forall (v, \omega) \in V_r$. Here, $V_r$ is the set of collision-free velocities ($\forall (v, \omega) \, \text{s.t} \, traj^{\mathcal{O}}(v, \omega) \bigcap obs_t = \emptyset$) that are reachable from the robot's current velocity based on its acceleration constraints. In equation \ref{eqn:dwa_obj_func}, $head(.)$, $obs(.)$, and $vel(.)$ are the cost functions~\cite{DWA} to quantify a velocity pair's (($v, \omega$) omitted on RHS for readability) heading towards the goal, distance to the closest obstacle in $obs_{t}$ from $traj^{\mathcal{O}}(v, \omega)$, and the forward linear velocity of the robot, respectively. $\sigma$ is a smoothing function and $\alpha_i, (i=1,2,3)$ are adjustable weights. 
\section{CoNVOI: Context-aware Navigation using VLMs in Outdoors and Indoors}

\ours \, uses two novel components: 1. a context-based prompting method to query a large VLM with an appropriate \textit{behavior} text prompt, and 2. a multi-modal visual marking method to prepare the large VLM's input image. Using both components \ours \, extracts a reference trajectory $\mathcal{T}_{ref}$ from the VLM. $\mathcal{T}_{ref}$ obeys the explicit and implicit rules associated with the robot's current context/scenario, and also encodes human-like preferences (e.g. opting to walk on pavement instead of grass) during navigation.

\subsection{Context-based Behavior Prompt}
VLMs utilize an image and text prompt inputs to output a text response. \ours's first component is an approach to frame a context-based behavioral text prompt for the VLM. To this end, we formulate a set of contexts/scenarios $\mathcal{C} = \{c_1, c_2, ..., c_k\}$ that the robot would operate in (e.g. indoor corridors, in the presence of humans, outdoor terrains, crosswalk, etc.), and a set of robot behaviors for each context. The behaviors $\mathcal{B} = \{b_1, b_2, ..., b_k\}$ are specified as simple text phrases such as ``keep close to the right wall" or ``move on pavement", etc. Our formulation allows for easily adding new scenarios and associated behaviors to $\mathcal{C}$ and $\mathcal{B}$, respectively.

Next, we query a light-weight VLM (e.g. CLIP \cite{clip}) with $I^{RGB}_t$ and text prompts $T_{i}$ framed as \textit{``This is a picture of $c_i$"} ($\forall i$), and obtain a set of probabilities $\{p_1, p_2, ... , p_k\} \, \big(\sum_i p_i = 1\big)$, where $p_i = P(c_i |I^{RGB}_t, \{T_{1}, T_{2}, ... , T_{k}\})$ is the probability that the context/scenario in $I^{RGB}_t$ corresponds to context $c_i$. Let $c_{max, t}$ correspond to the context with maximum probability at time $t$, and $b_{max}$ be its associated robot behavior. Next, we frame a behavioral prompt $T_{b_{max}}$ in the format \textit{``You are navigating a robot in this scenario. The numbers marked in the image denote regions where you can navigate the robot. Pick a list of numbers marked in the image such that the robot \rule{1cm}{0.15mm}"}, and the blank is filled with the behavioral phrase in $b_{max}$.

\subsection{Multi-modal Visual Marking}
To prepare the image input for the VLM, we take into account the following considerations. Firstly, VLMs do not reliably infer spatial relationships, such as the distance of a point from the camera and the relative locations of various regions in the image (e.g., \textit{point A is closer to the right side of the image})\cite{liu2023vlm_spatial}. Secondly, the VLM does not need to \textit{focus} on obstacle regions since its task is solely to identify regions within the free space to identify \textit{how} the robot should navigate. To this end, we propose to use the correlation between the robot's occupancy grid map $\mathcal{O}_t$ and the RGB image $I^{RGB}_t$ to detect the free space in the robot's vicinity, and overlay markers (numbers) \cite{pivot} on the corresponding free space in $I^{RGB}_t$. This approach combines the benefits of the spatial information encoded in occupancy grids with the semantic information in RGB images and results in a richer input to the VLM. 

\subsubsection{Detecting Free Regions to Add Markers} \label{sec:checking-free-space}
We consider $l$ rows and $m$ columns of grids each on $\mathcal{O}_t$ as candidate locations where the robot could traverse to exhibit context-aware behaviors. These grids are uniformly spaced by a distance $d_{col}$ along the columns, and $d_{row}$ along the rows. Let us consider the $ij^{th}$ candidate grid $\mathbf{g}_{ij}, \, i \in \{1, 2, ..., l\}, j \in \{1, 2, ..., m\}$. Let $L_{ij}$ be the line connecting $\mathbf{g}_{ij}$ and the robot's location $(n/2, n/2)$ in $\mathcal{O}_t$. $\mathbf{g}_{ij}$ is considered as a navigable candidate if,

\begin{equation}
    L_{ij} \bigcap obs_{t} = \emptyset. 
    \label{eqn:marker-in-obstacle}
\end{equation}

This ensures that locations obstructed by obstacles from the robot and its camera's view are not considered viable for context-aware navigation. We construct the set of candidate coordinates satisfying equation \ref{eqn:marker-in-obstacle} as $X^{\mathcal{O}}_t = \{\mathbf{g}_{ij}\} \, \forall (i, j)$. The points in $X^{\mathcal{O}}_t$ are transformed into the RGB frame as, 

\begin{equation}
    (x^{RGB}_{ij}, y^{RGB}_{ij}) = T^{RGB}_{\mathcal{O}} \cdot \mathbf{g}_{ij} \quad \forall \mathbf{g}_{ij} \in X^{\mathcal{O}}_t.
    \label{eqn:marker-pixel-coordinates}
\end{equation}

\no We construct the set of candidate coordinates as $X^{RGB}_t = \{(x^{RGB}_{ij}, y^{RGB}_{ij})\} \, \forall (i, j)$. 
The points in $X^{RGB}_t$ are ordered from left to right and row by row (see Fig. \ref{fig:system-arch}). Next, we mark numbers on all the points in $X^{RGB}_t$ in ascending order on $I^{RGB}_t$ and refer to this marked image as $I^{Mark}_t$.

\subsection{Prompting the Large VLM}
Finally, using $T_{b_{max}}$ and $I^{Mark}_t$, we prompt a large VLM \cite{gpt-4v,gemini} to extract a reference path as,

\begin{equation}
    \mathcal{M}_{ref} = \text{VLM}(I^{Mark}_t, T_{b_{max}}). 
    \label{eqn:vlm-query}
\end{equation}

\no Here, $\mathcal{M}_{ref}$ is a list of numbers (markers) marked on $I^{Mark}_t$. Their corresponding set of grid locations relative to $\mathcal{O}_t$ are obtained by using the numbers in $\mathcal{M}_{ref}$ as indices since they are marked orderly in $I^{Mark}_t$. Hence, the reference path in the occupancy map frame is $\mathcal{T}^{\mathcal{O}}_{ref} = X^{\mathcal{O}}_t(\mathcal{M}_{ref})$. Their locations in the robot frame can be calculated as $\mathcal{T}^{rob}_{ref} = T^{rob}_{\mathcal{O}} \cdot \mathcal{T}^{\mathcal{O}}_{ref}$.

\subsection{Reference Path Following} 
When the robot follows the reference path $\mathcal{T}^{rob}_{ref} = \{(x^{rob}_{ref,1}, y^{rob}_{ref,1}), (x^{rob}_{ref,2}, y^{rob}_{ref,2}), ... , (x^{rob}_{ref,last}, y^{rob}_{ref,last}) \} \,\, i = \{1, 2, ... , last\}$, it exhibits the most appropriate navigation behaviors for the context it encounters. These coordinates are repeatedly transformed into the robot's coordinate frame as the robot moves. To follow the reference path, we modify the motion planner \cite{DWA} explained in section \ref{sec:planner-background} as follows. Let $\theta_{goal}$ be the angle between the robot's current heading direction and the vector pointing to its goal direction. We first formulate a reference path following cost as,

\begin{equation}
 \begin{split}
    ref(v, \omega) = |y^{rob}_{ref,i} - y^{rob}_{t + t_{hor}}|, \,\, y^{rob}_{t + t_{hor}} \in traj^{rob}(v, \omega), \\ 
    \text{and } i \,\,  \text{s.t} \,\, x^{rob}_{ref,i} > 0 \,\, \text{and} \\ \,\, x^{rob}_{ref,i} = min(x^{rob}_{ref,1}, x^{rob}_{ref,2}, ... , x^{rob}_{ref,last}),
\end{split}  
\label{eqn:ref-path-cost}
\end{equation}

\no Equation \ref{eqn:ref-path-cost} ensures that the robot picks $(v, \omega)$ trajectories that minimize the lateral distance between the robot and a reference path point that is immediately ahead of the robot. A new objective function for planning as,

\begin{equation} 
Q_2(v, \omega) = 
    \begin{cases}
    Q_1(.) + \alpha_4 \cdot ref(.) \,\, & \text{if} \, |\theta_{goal}| \le \theta_{fov}, \\
    Q_1(.) & \text{Otherwise}.  
    \end{cases}
    \label{eqn:new-obj-func}
\end{equation}

\no Here, $\theta_{fov}$ is the field of view of the robot's camera. This condition helps to ensure that the robot does not stray away from the goal direction by following the VLM's reference paths.

\subsection{Reference Path Extrapolation and Re-querying}
Since large VLMs require several seconds to respond to queries (such as in equation \ref{eqn:vlm-query}), continuously querying to obtain new reference paths is computationally infeasible for real-time navigation. Therefore, \ours \, reduces the frequency of querying the large VLM by extrapolating the current reference path, if the context has not changed.  

Once the robot obtains a reference path and starts following it, \ours \, repeatedly queries CLIP to check for context changes. If $c_{max, t} = c_{max, t'} (t' > t)$, and $\text{dist}((x^{rob}_{ref,last}, y^{rob}_{ref,last})) < d_{thresh}$ (since $(0, 0)$ is the robot's location in its own frame, and $d_{thresh}$ is a distance threshold), we linearly extrapolate the current reference path by fitting a line to the points in $\mathcal{T}^{rob}_{ref}$ and extending it further to a point $(x^{rob}_{ref,e}, y^{rob}_{ref,e})$ by distance $d_{row}$. Next, we check if the point $(x^{\mathcal{O}}_{ref,e}, y^{\mathcal{O}}_{ref,e})$ is obstacle-free (similar to section \ref{sec:checking-free-space}). If it is, we add the point to the RGB image at $(x^{RGB}_{ref,e}, y^{RGB}_{ref,e})$, and crop a $n_{pat} \times n_{pat}$ image patch $I^{pat}$ (see Fig. \ref{fig:system-arch}) to query CLIP if the patch is an image of a paved (indoors or sidewalks) or unpaved (grass, gravel, asphalt) surface. If $CLIP(I^{pat})$ is paved, $(x^{rob}_{ref,e}, y^{rob}_{ref,e})$ is used as the next waypoint for the planner to follow. Otherwise, the large VLM is queried for the next reference path.

If the context has changed, \ours \, queries the large VLM once $\text{dist}((x^{rob}_{ref,last}, y^{rob}_{ref,last})) < d_{thresh}$. $d_{thresh}$ is chosen such that $d_{thresh} \approx v_{max} \cdot t_{query}$, where $v_{max}$ is the maximum linear velocity of the robot and $t_{query}$ is the average time for a large VLM to respond to a query.

\section{Results and Evaluations}
In this section, we explain \ours's implementation on real robots, and analyze and evaluate its performance through comparisons and ablations. 

\begin{figure*}[t]
    \centering
    \includegraphics[width=0.9\linewidth]{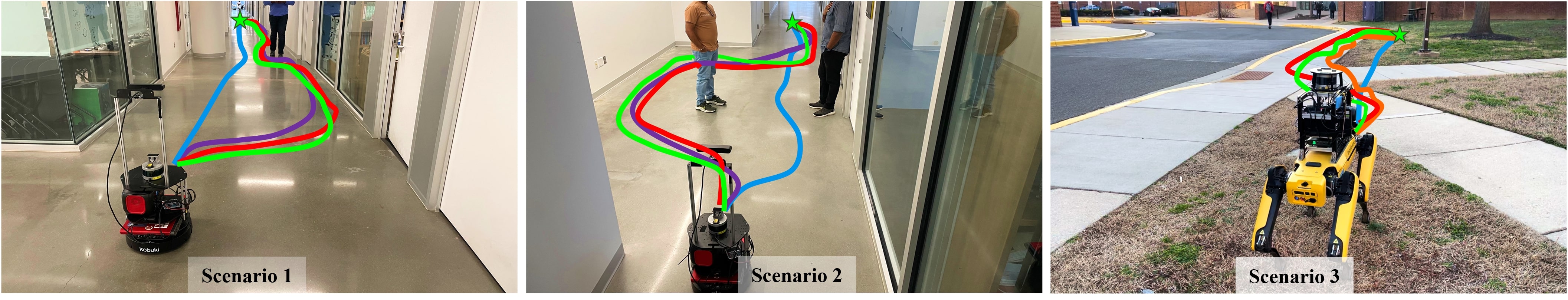}
    \caption{\small{Robot trajectories when navigating in different complex indoor and outdoor environments using various methods: \ours \, (in green), teleoperated by a human (in red), DWA \cite{DWA} (in blue), Frozone \cite{frozone} (in violet), GA-Nav \cite{guan2021ganav} (in orange). \ours \, exhibits social-compliant behaviors such as not moving in-between humans even if there is sufficient space, similar to Frozone, a method formulated for indoor social navigation. \ours's behaviors also match GA-Nav, a semantic segmentation-based navigation approach that prefers to navigate on smooth, well-paved outdoor terrains. \ours \, achieves these behaviors in a zero-shot manner, and does not require domain-specific fine-tuning. \ours's trajectories also closely match human-teleoperated ground truth paths (in red).}}
    \label{fig:robot-traj}
\end{figure*}

\begin{figure}[t]
    \centering
    \includegraphics[width=\columnwidth]{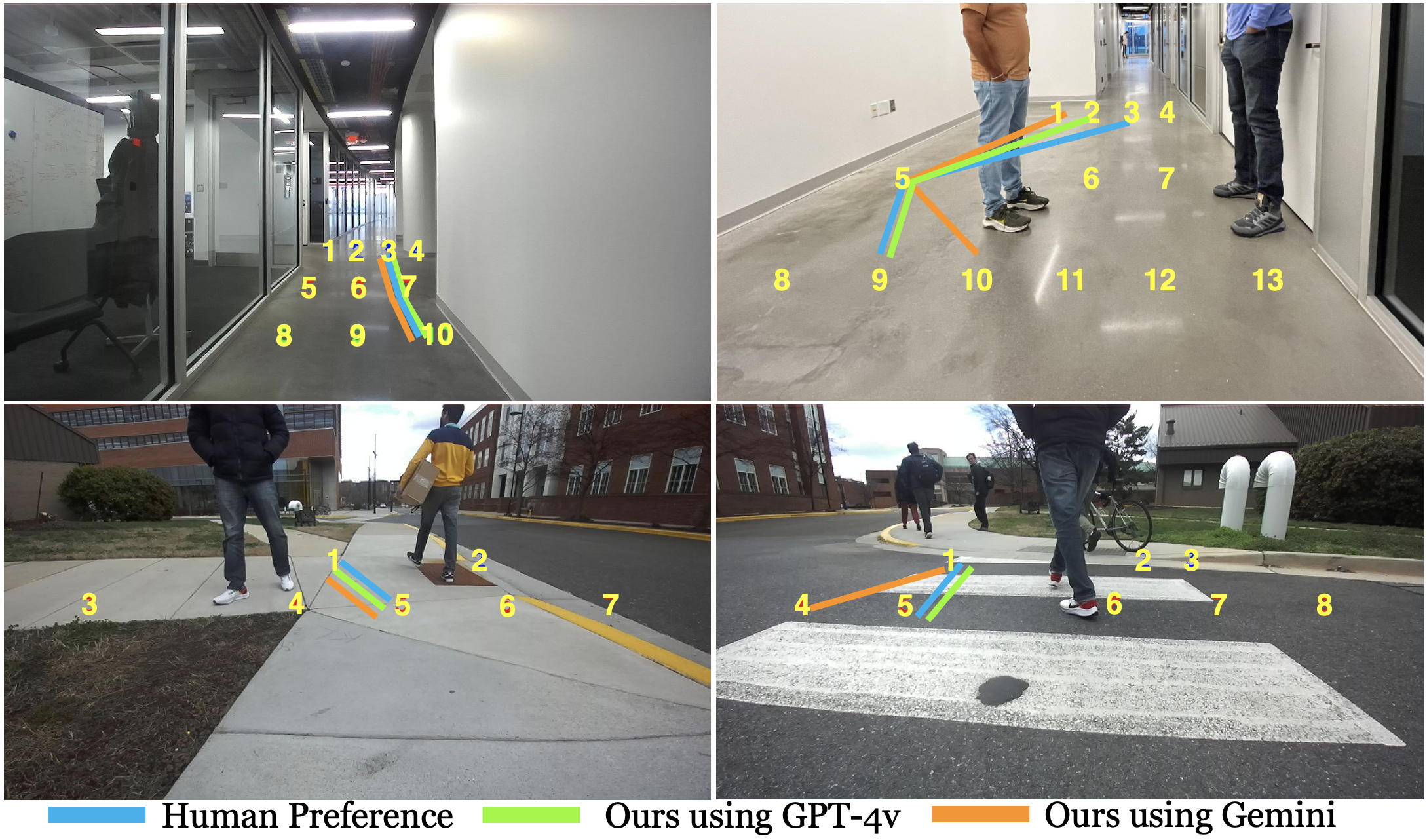}
    \caption{\small{Qualitative comparison of the reference trajectory generated by GPT-4v \cite{gpt-4v} (in green), and Gemini \cite{gemini} (in orange) compared with human provided ground truth (in blue) by connecting the marked numbers (in yellow). We observe that the reference paths generated by the VLMs is comparable to that of the human-preferred path in many complex indoor and outdoor environments.}}
    \label{fig:ref-path-comp}
    \vspace{-15pt}
\end{figure}

\subsection{Implementation}
\ours \, is implemented on a Turtlebot 2 for indoor evaluation, and on a Boston Dynamics Spot robot for outdoor evaluation. Both robots are equipped with a Velodyne VLP16 lidar, and a Zed 2i camera. \ours \, is executed on a laptop with an Intel i7 11th generation CPU and an Nvidia RTX 3060 GPU. For context detection, CLIP is executed locally on the laptop, and for the reference path queries, we use the online API for GPT-4v \cite{gpt-4v}, and Gemini \cite{gemini}. We use the following values for the parameters in our formulation: $n = 200, \alpha_1 = 10, \alpha_2 = 7, \alpha_3 = 1, \alpha_4 = 7.5, l = 2 \, \text{for outdoors}, 3 \, \text{for indoors}, m = 6, d_{row} = 2.5m, d_{col} = 2m, k = 5, v_{max} = 0.3 \text{m/s}, t_{query} = 5s, d_{thresh} = 4m, n_{pat} = 200$. 

\subsection{Comparison Methods}
We compare \ours \, with several existing methods for indoor and outdoor navigation. They are: \\
\begin{itemize}
    \item \textbf{DWA} \cite{DWA}: The dynamic window approach, a baseline motion planner that performs simple collision avoidance and goal-reaching behaviors.

    \item \textbf{GA-Nav} \cite{guan2021ganav}: A planner that uses semantic segmentation-based outdoor terrain classification to assign traversability costs to various terrains and computes trajectories on smooth (low-cost) terrains.

    \item \textbf{PSP-Net} \cite{pspnet}: A planner similar to GA-Nav but employing PSP-Net for semantic segmentation.

    \item \textbf{Frozone} \cite{frozone}: A social-compliant indoor navigation method that computes trajectories that are least obtrusive to surrounding pedestrians. We enhance Frozone's formulation by incorporating a method to detect corridors from the robot's local occupancy grid. Additionally, we assign higher costs to regions closer to the left wall, which encourages the robot to stay closer to the right side.
\end{itemize}

In addition to baseline methods, we conduct thorough ablation studies by analyzing our approach's reference path selection performance by removing its multi-modal visual marking, and context-based prompting components.


\subsection{Evaluation Metrics}
We use the following quantitative metrics to evaluate the navigation performance of all the methods in real-world scenarios.
\begin{itemize}
    \item \textbf{Fréchet Distance w.r.t. Human Teleoperation}: Measures the Fréchet distance \cite{alt1995frechet_distance} (a measure of similarity between two curves) between a human teleoperated robot trajectory versus a comparison method's trajectory.
    \item \textbf{Normalized Trajectory Length}: The robot's trajectory length normalized over the straight-line distance to the goal.
    \item \textbf{Mean Velocity}: The robot's linear velocity (that indicates its progress towards the goal) averaged over all the trials. 
\end{itemize}


\no We use the following metrics to evaluate \ours's reference path selecting capabilities by comparing it with a human chosen ground truth points in the marked image. 
\begin{itemize}
    \item \textbf{Reference Path Error}: Measures the distance between the mean points in the ground truth path and the VLM's output reference path.
    \item \textbf{Cosine Similarity}: Quantifies the alignment between the ground truth and the VLM's reference path in terms of orientation.
    \item \textbf{Number of unacceptable paths}: Counts the number of times a reference path was on an unacceptable terrain/location (e.g. on an obstacle, asphalt road that could have traffic, etc).
    \item \textbf{Inference Time}: The time between sending a query to the large VLM and it returning a reference path. This includes network latencies.
\end{itemize}


\subsection{Navigation Testing Scenarios}
We compare all the methods across four types of scenarios: 1. Indoor corridors, 2. Indoor scenarios with people, 3. Outdoor scenarios with multiple terrains, 4. Outdoor crosswalks. Additionally, we illustrate the addition of behaviors for a new scenario and showcase the resulting robot behavior.

\begin{table}
\resizebox{\columnwidth}{!}{%
\begin{tabular}{ |c |c |c |c |c |c|} 
\hline
\textbf{Scenario} & \textbf{Method} & \multicolumn{1}{|p{1.5cm}|}{\centering \textbf{Reference \\ Path Error} $\downarrow$} & \multicolumn{1}{|p{1.5cm}|}{\centering \textbf{Cosine \\ Similarity $\uparrow$}} & \multicolumn{1}{|p{1.5cm}|}{\centering \textbf{$\%$ Unacceptable \\Paths $\downarrow$}}& \multicolumn{1}{|p{1cm}|}{\centering \textbf{Inference \\Time $\downarrow$} } 
\\ [0.5ex] 
\hline
\multirow{6}{*}{\rotatebox[origin=c]{0}{\makecell{\textbf{Scen.}\\\textbf{1}}}} 
& Gemini w/o MMVM w. CbP & 1.442 & 0.926 & 0.857 & 5.24 \\
& GPT4 w/o MMVM w. CbP & 1.576 & 0.913 & 0.913 & 4.86 \\
& Gemini w. MMVM w/o CbP  & 0.315 &  0.995 & \textbf{0.0} & 4.78 \\
& GPT4 w. MMVM w/o CbP  & 0.335 & 0.993 & \textbf{0.0} & 4.10 \\
& Gemini w. \ours  & 0.202 & \textbf{0.996} & \textbf{0.0} & 3.76 \\
& GPT4 w. \ours  & \textbf{0.187} & 0.995 & \textbf{0.0} & \textbf{3.48} \\
\hline
\multirow{6}{*}{\rotatebox[origin=c]{0}{\makecell{\textbf{Scen.}\\\textbf{2}}}} 
& Gemini w/o MMVM w. CbP & 1.425 & 0.914 & 0.786 & 5.11 \\
& GPT4 w/o MMVM w. CbP & 2.325 & 0.802 & 0.798 & 4.69 \\
& Gemini w. MMVM w/o CbP  & 0.484 & 0.989 & 0.278 & 4.91 \\
& GPT4 w. MMVM w/o CbP  & 0.589 & 0.988 & 0.223 & 4.46 \\
& Gemini w. \ours  & \textbf{0.444} & 0.981 & \textbf{0.111} & 3.55 \\
& GPT4 w. \ours  & 0.456 & \textbf{0.991} & 0.167 & \textbf{3.28} \\
\hline

\multirow{6}{*}{\rotatebox[origin=c]{0}{\makecell{\textbf{Scen.}\\\textbf{3}}}} 
& Gemini w/o MMVM w. CbP & 0.891 & 0.973 & 0.433 & 6.35 \\
& GPT4 w/o MMVM w. CbP & 0.865 & 0.977 & 0.366 & 5.60 \\
& Gemini w. MMVM w/o CbP  & 0.817 & \textbf{0.978} & 0.291 & 6.28 \\
& GPT4 w. MMVM w/o CbP  & 0.872 & 0.971 & 0.125 & 5.46 \\
& Gemini w. \ours  &  0.979 & 0.970 & 0.125 & 4.88 \\
& GPT4 w. \ours  & \textbf{0.785} & 0.975 & \textbf{0.083} & \textbf{4.15} \\
\hline

\multirow{6}{*}{\rotatebox[origin=c]{0}{\makecell{\textbf{Scen.}\\\textbf{4}}}} 
& Gemini w/o MMVM w. CbP & 1.210 & 0.957 & 0.285 & 6.36 \\
& GPT4 w/o MMVM w. CbP & 1.205 & 0.966 & 0.190 & 4.89 \\
& Gemini w. MMVM w/o CbP  & 1.229 & 0.938 & 0.223 & 5.98 \\
& GPT4 w. MMVM w/o CbP  & 1.159 & 0.962 & 0.166 & 4.95 \\
& Gemini w. \ours  & 1.205 & 0.977 & \textbf{0.055} & 5.26 \\
& GPT4 w. \ours  & \textbf{0.819} & \textbf{0.978} & \textbf{0.055} & \textbf{4.83} \\
\hline
\end{tabular}
}
\caption{\small{Ablation studies comparing the impact of removing multi-modal visual marking (MMVM) and context-based prompting (CbP) from our formulation on generating reference paths. We also compare the use of two state-of-the-art large VLMs: GPT-4v \cite{gpt-4v} and Gemini \cite{gemini}. Our results show that MMVM and CbP significantly improve the VLM's performance in generating human-like reference paths.}
}
\label{tab:ablation}
\end{table}

\begin{table}[t]
\centering
\resizebox{\columnwidth}{!}{%
\begin{tabular}{ |c |c |c |c |c |} 
\hline
\textbf{Scenario} & \textbf{Method} & \multicolumn{1}{|p{1.5cm}|}{\centering \textbf{Fréchet Dist. $\downarrow$}} & \multicolumn{1}{|p{1.5cm}|}{\centering \textbf{Norm. \\ Traj.} \\ \textbf{Length} $\approx 1$} & \multicolumn{1}{|p{1cm}|}{\centering \textbf{Mean\\Velocity $\uparrow$} } 
\\ [0.5ex] 
\hline
\multirow{4}{*}{\rotatebox[origin=c]{0}{\makecell{\textbf{Scen.}\\\textbf{1}}}} 
 & DWA\cite{DWA} & 0.864& 1.082&  0.395 \\
 & Frozone\cite{frozone} & 0.577& 1.239& 0.392 \\
 & CoNVOI (ours)  & 0.521& 1.207&  0.289\\
\hline
\multirow{4}{*}{\rotatebox[origin=c]{0}{\makecell{\textbf{Scen.}\\\textbf{2}}}} 
 & DWA\cite{DWA} & 1.087& 1.105&   0.390\\
 & Frozone\cite{frozone} & 0.866& 1.187& 0.388 \\
 & CoNVOI (ours)  & 0.709& 1.194&  0.276\\
\hline

\multirow{4}{*}{\rotatebox[origin=c]{0}{\makecell{\textbf{Scen.}\\\textbf{3}}}} 
 & DWA\cite{DWA} & 2.475& 1.067&  0.485 \\
 & PSP-Net \cite{pspnet} & 2.188 & 1.121 & 0.458 \\
 & GA-Nav\cite{guan2021ganav}  & 1.957 & 1.160 & 0.473\\ 
 & CoNVOI (ours)  & 1.083& 1.213& 0.238 \\
\hline

\multirow{4}{*}{\rotatebox[origin=c]{0}{\makecell{\textbf{Scen.}\\\textbf{4}}}} 
  & DWA\cite{DWA} & 2.095& 1.091&   0.491\\
 & PSP-Net \cite{pspnet} & 2.068 & 1.089 & 0.455 \\
 & GA-Nav\cite{guan2021ganav}  & 2.157& 1.112& 0.466\\ 
 & CoNVOI (ours)  & 0.863& 1.237& 0.243\\
\hline
\end{tabular}
}
\caption{\small{Performance comparisons when a robot uses \ours \, DWA motion planner \cite{DWA} that performs only obstacle avoidance and goal-reaching, Frozone \cite{frozone}, a socially-compliant indoor navigation method, and PSP-Net \cite{pspnet} and GA-Nav \cite{guan2021ganav}, two semantic segmentation-based outdoor navigation methods. \ours \, generates trajectories that are most similar to human teleoperated trajectories (lowest Fréchet distance). Qualitatively, \ours \, performs similar to social and outdoor navigation methods in a zero-shot manner and is applicable to both indoor and outdoor environments.}
}
\label{tab:comparison_table}
\vspace{-15pt}
\end{table}

\subsection{Analysis and Comparisons}
The results of our qualitative and quantitative comparisons are shown in Fig. \ref{fig:robot-traj}, and Table \ref{tab:comparison_table}, respectively. The results of our ablation studies are shown in Table \ref{tab:ablation}. 

\textbf{Ablation Study:} We analyze the effects of our multi-modal visual marking (MMVM), and context-based prompting on generating the reference path by removing these components individually from \ours. Next, we compare the generated reference path with human-provided reference paths (see Fig. \ref{fig:ref-path-comp}). First, removing MMVM, the marked image $I^{Mark}_t$ would contain numbers on obstacles (walls, people, etc.). Therefore, while choosing the numbers for the reference path, the VLM must also ensure that they do not lie on obstacles (thus being responsible for avoiding them). However, we observe that this results in high reference path errors, and $\%$ of unacceptable paths as a subset of the chosen points often lie on obstacles in indoor scenarios (1 and 2). Additionally, the cosine similarity when compared with human GT is low, indicating that the paths are not generally in the direction preferred by a human. In outdoor environments, removing MMVM does not affect the metrics as adversely, since these scenarios are expansive and less constrained by humans. However, the $\%$ of unacceptable paths is higher as the returned reference paths some times intersect with walking pedestrian obstacles. The inference rate is also increased since the VLM must now pay attention to more regions before choosing points for the reference path. We observe that current VLMs cannot be used for zero-shot obstacle avoidance reliably. 

In the second ablation study, we removed context-based prompting (CbP) and used a single detailed prompt describing the rules for navigating all associated scenarios. We observe that this leads to a performance comparable to \ours \, in terms of the path error and cosine similarity. However, there is a considerable increase in the VLM's inference time. This could be due to the higher number of tokens the VLM must process given a detailed text prompt. Additionally, we observe a significant increase in the $\%$ of unacceptable paths in scenarios 2 and 4, where certain intricate behaviors (avoid moving in-between people, navigate on crosswalk) are expected. Such behaviors may be harder to interpret accurately when included with behaviors for other scenarios in the detailed prompt. We observe the lowest path errors, $\%$ of unacceptable paths, inference time, and highest cosine similarity for \ours \, (with MMVM, CbP) in all cases. Since using \ours \, with GPT-4v \cite{gpt-4v} has the least inference rate along with the best reference paths, we use it for comparison with other navigation methods in Table \ref{tab:comparison_table}.

\textbf{Comparison with Navigation Methods:} We compare all the navigation methods relative to human teleoperation, which is considered the \textit{best} trajectory the robot could take in its current context. We observe that \ours's zero-shot path resemble the human teleoperated path the closest (lowest Fréchet distance in Table \ref{tab:comparison_table}) in all the scenarios (see Fig. \ref{fig:cover_image}, and \ref{fig:robot-traj}). 

We observe that \ours \, typically leads to a higher normalized trajectory length due to higher deviations from the goal to follow the context-based navigation behavior, similar to the human teleoperated path. Additionally, \ours \, has a low mean velocity due to its limiting maximum velocity ($v_{max} = 0.3$ m/s) set to ensure uninterrupted navigation, due to the high inference time current VLMs require. However, we highlight that \ours \, demonstrates all these context-based behaviors in a zero-shot manner, utilizing simple text and visual prompts. Unlike existing model-based and deep learning-based methods, it has demonstrated wide applicability across various indoor and outdoor environments. 

\textbf{Handling Novel Scenarios:} We also highlight that new types of scenarios can be easily added to \ours \, without any reformulation. We add a scenario with a detour sign and instruct \ours \, to deviate the robot in the direction of the detour sign's arrow. We observe that \ours \, immediately adapts to the new scenario, and navigates to its goal behind the sign (Fig. \ref{fig:cover_image} [bottom]). This signifies how \ours \, can use VLMs' complex visual understanding and reasoning capabilities for navigation in the real-world. 

\textbf{Extrapolation Benefits:} We observe that \ours's extrapolation helps avoid unnecessarily requerying the large VLM in environments where the context does not change until the goal (e.g. in corridors, straight pavements). On average, extrapolation helps reduce the requerying by $\sim 50\%$ in such environments. In dynamic scenarios with walking humans, in the presence of several terrains (extrapolated point could lie in unpaved terrain), the VLM is requeried to guide the robot. For more results, we point the reader to our technical report.

\textbf{VLM Hallucinations:} Since during navigation the large VLM is typically fed with a sequence of \textit{similar}-looking images and text prompts, we observe that it often starts to hallucinate and return reference path numbers in a certain pattern (e.g. [1, 2, 3, 4, 5], or [1, 2, 1, 2]) without considering the scenario. This occurs when the VLM uses its recent memory to perform the current task. To alleviate hallucinations, we add a phrase to the text prompt to instruct the VLM that it must consider the current query as a new task and disregard past inputs and outputs.  

\textbf{Wavy Trajectories}: We observe that when there are several correct paths that satisfy the behavior specified in the context-based text prompt, the VLM may return any one of them. This could sometimes result in a wavy trajectory (as shown in Fig. \ref{fig:cover_image} on the crosswalk, while passing the barricade). This is primarily due to the spacing parameter $d_{col}$ between the numbers marked in the image, which affects the spatial resolution of the reference paths. However, decreasing $d_{col}$ leads to more numbers annotated on the image, which leads to higher inference rates to process, similar to removing MMVM in Table \ref{tab:ablation}. Such limitations could be alleviated with faster large VLMs in the future.  

\section{Conclusions, Limitations, and Future Work}
We present \ours, a novel method to use compact and large VLMs for indoor and outdoor navigation without any additional training or fine-tuning. \ours \, utilizes a novel multi-modal visual marking scheme, and a context-based prompting method to achieve complex behaviors (e.g. avoiding moving in-between groups of humans, interpreting signs, etc) in various indoor and outdoor scenarios. \ours \, also uses an extrapolation method to avoid unnecessary requerying of the large VLM that reduces the queries by up to $50\%$ in certain scenarios.

Our method has a few limitations. Since \ours \, uses a large VLM that is hosted remotely, its processing time and network latency affects the robot's navigation and its maximum velocity. In outdoor environments, since network speeds also depend on weather conditions, we observe some rare instances where the robot does not receive a reference path, and uses the baseline planner to navigate. Additionally, this limits VLMs' use in highly dynamic environments with crowds. However, such limitations can be overcome as large VLMs become faster, and locally executable. We also observe that outputs could be sensitive to vague terms in the text prompt (e.g. ``\textit{cross the street like a human}" instead of specifying ``\textit{stay on the crosswalk}"). Our current formulation does not use any global information like a top-down view of the environment, which would be needed to assess factors like where to cross the road, when the crosswalk is not visible from the robot's camera. We plan to address these limitations in our future work, and extend our formulation to long-range navigation.   

\bibliographystyle{IEEEtran}
\bibliography{References}
\end{document}